# OHRID-Retail: An Open Multimodal Dataset of Human Activity in Retail Environments

Xiangrui Wang[1]†, Yuetong Wu[1]†, Jalen Beeman[2], Robert Cook[3], Yu Gu[3], Nathanial Pearson[2], Trevor Smith[3], Read Hayes[4], Boyi Hu[1]*

[1]Department of Industrial and Systems Engineering, University of Florida, Gainesville, FL, USA.

[2]Lane Department of Computer Science and Electrical Engineering, West Virginia University, Morgantown, WV, USA

[3]Department of Mechanical, Materials and Aerospace Engineering, West Virginia University, Morgantown, WV, USA.

[4]Loss Prevention Research Council (LPRC), University of Florida, Gainesville, FL, USA

*Corresponding author E-mail: boyihu@ise.ufl.edu

†These authors contributed equally to this work.

## Abstract

Open datasets describing human behavior in environments shared with mobile robots remain limited, particularly for retail activities that combine locomotion, reaching, object handling, and robot-guided movement. This paper introduces OHRID-Retail, an open, human-centered multimodal dataset collected from 16 healthy adults performing a simulated shelf-picking task under three within-participant conditions: no robot, low-speed robot guidance, and high-speed robot guidance. Each participant completed two trials per condition. Whole-body kinematics were recorded using 17 Xsens Awinda inertial sensors and muscle activity was measured at 10 locations using Delsys Trigno surface electromyography sensors. Descriptive analyses demonstrate variation in whole-body movement intensity and muscle activation across robot-interaction conditions and body locations. OHRID-Retail provides openly available raw recordings, processed measures, documentation, and reproducible analysis resources. The dataset can support research in human activity recognition, multimodal sensor fusion, occupational biomechanics, ergonomics, human-aware robot navigation, and human–robot interaction in retail and related shared environments.

## I. Introduction

Service robots are increasingly being introduced into retail stores, workplaces, healthcare facilities, and other human-populated environments, where they are expected to operate alongside people while performing tasks such as inventory monitoring, cleaning, transportation, security, and customer assistance. The growing presence of robots in these shared environments has also increased the need for data that capture the complexity and diversity of real-world operation. In recent years, large open robotics datasets have provided extensive observations of robot actions, perception, manipulation, and interactions with objects and physical environments. These resources have become important infrastructure for developing and evaluating robotic systems across diverse tasks and settings.

While existing robotics datasets provide increasingly rich information about the robot side of shared environments, understanding the human side is equally important for the development of human-aware robotic systems. People continuously move through shared spaces, change body positions, interact with objects, and respond to features of the surrounding environment. These behaviors create dynamic conditions that robotic systems must eventually perceive, interpret, and accommodate. Human-centered data describing activities and movement patterns can therefore complement robot-centered datasets and support research on how robots may safely and effectively operate in spaces designed primarily around human activities.

However, open datasets capturing naturalistic or semi-naturalistic human behavior in environments relevant to future human–robot coexistence remain comparatively limited. Retail environments provide a particularly useful context for addressing this gap because they combine structured physical spaces with a diverse range of human activities. Within a retail environment, individuals may walk through aisles, stop and change direction, browse shelves, reach for and handle products, bend or squat, carry objects, and perform other activities that produce substantial variation in movement and interaction with the surrounding environment. Systematically capturing these behaviors provides a human-centered complement to large robot-centric open-data efforts and establishes a data resource that can support future studies of human-aware robotics and human–robot interaction.

To address this need, we introduce OHRID-Retail, an open multimodal dataset of human activity in retail environments developed as part of the Open Human–Robot Interaction Data Initiative for Retail Applications**.** The dataset includes 16 participants performing a structured shelf-picking task under three within-participant conditions: no robot (NR), low-speed robot guidance (LOW), and high-speed robot guidance (HIGH). Data were collected in a controlled retail-like mixed-reality environment containing a shelving unit, numbered objects, a collection bin, projected safe and unsafe working regions, and an autonomous mobile robot. Human activity was recorded using 17 Xsens Awinda inertial sensors for whole-body kinematics and 10 Delsys Trigno surface electromyography sensors for muscle activity. The experimental platform also used Vicon spatial tracking and robot-state information to support real-time hazard detection and robot guidance. The open release provides the collected data together with documentation describing the experimental environment, activity protocols, sensing configuration, data organization, and processing procedures. The current release focuses primarily on human activities and movement, providing a human-side data resource that can serve as a foundation for future expansion toward explicit human–robot interaction data.

The primary contributions of OHRID-Retail are threefold. First, we provide an open human-centered dataset capturing human activities and movement within a retail environment relevant to future human–robot coexistence. Second, we provide a standardized multimodal data-collection and organization framework that facilitates access, interpretation, and reuse of the collected data. Third, we provide a descriptive characterization of the dataset, including its composition, coverage, and behavioral diversity across participants and retail activities. Together with the accompanying open-access website and data repository, OHRID-Retail is intended to provide a reusable data resource for research involving human activity, human movement, ergonomics, human-aware robotics, and human–robot interaction in retail and related shared environments.

## II. Related Work

### A. Human Activity and Motion Datasets

Open human activity and motion datasets can be grouped by sensing configuration and research focus. First, wearable human activity datasets, such as PAMAP2 and Opportunity++, provide inertial, physiological, and environmental sensor streams for recognizing locomotion, gestures, and activities of daily living (Ciliberto et al., 2021; Reiss & Stricker, 2012). Second, vision and skeleton-centered datasets, including NTU RGB+D and MMAct, support action recognition using RGB video, depth, body keypoints, and wearable sensing modalities (Kong et al., 2019; Shahroudy et al., 2016). Third, high-fidelity motion datasets, such as HumanEva, MoVi, RELI11D, and ActionSense, provide synchronized motion-capture or multimodal recordings (DelPreto et al., 2022; Ghorbani et al., 2021; Sigal et al., 2010; Yan et al., 2024). ActionSense additionally includes muscle activity, eye tracking, force/tactile sensing, and external cameras during kitchen tasks (DelPreto et al., 2022). Finally, large-scale video datasets such as Motion-X++ and Ego4D prioritize motion reconstruction and action understanding across daily-life settings (Grauman et al., 2022; Zhang et al., 2025).

### B. Robotics and Human–Robot Interaction Datasets

Large robot-learning datasets are commonly organized around robot observations, actions, and demonstrations. For example, DROID provides large-scale robot-manipulation trajectories collected across diverse real-world environments to train and evaluate generalizable robot policies (Khazatsky et al., 2024). HRI datasets additionally capture human and environmental signals in specific task contexts. HARMONIC records multimodal human, robot, and environmental data during shared-autonomy assistive eating, whereas HRI30 provides task-specific action videos for industrial human-robot collaboration (Newman et al., 2022; Iodice et al., 2022). Kaiwu further combines electromyography, eye tracking, hand sensing, and optical three-dimensional motion capture during assembly tasks for robot learning and human-robot collaboration (Jiang et al., 2025).

The present dataset (OHRID-Retail) differs in purpose and scope. Although the experimental protocol includes a retail mobile robot, the dataset is human-centered rather than robot-centric. It documents synchronized full-body inertial motion and surface electromyography during a structured retail-relevant task under different levels of robot assistance. The controlled within-participant design supports analyses of how robot presence and motion speed relate to human movement and physiological responses. Together with openly available raw recordings, processed features, and reproducible analysis resources, the dataset provides a resource for studying human performance in future retail environments where people and robots coexist.

## III. Data Collection Setup

### A. Experimental Environment

Data were collected in a controlled, retail-like mixed-reality environment designed to support repeatable shelf-picking tasks and close-proximity interactions between a participant and an autonomous mobile robot (AMR). The task area contained a shelving unit holding ten numbered boxes, a fixed participant starting location, a designated collection bin positioned to the right of the shelving unit, and sufficient open floor space for participant movement and robot navigation (Figure 1).

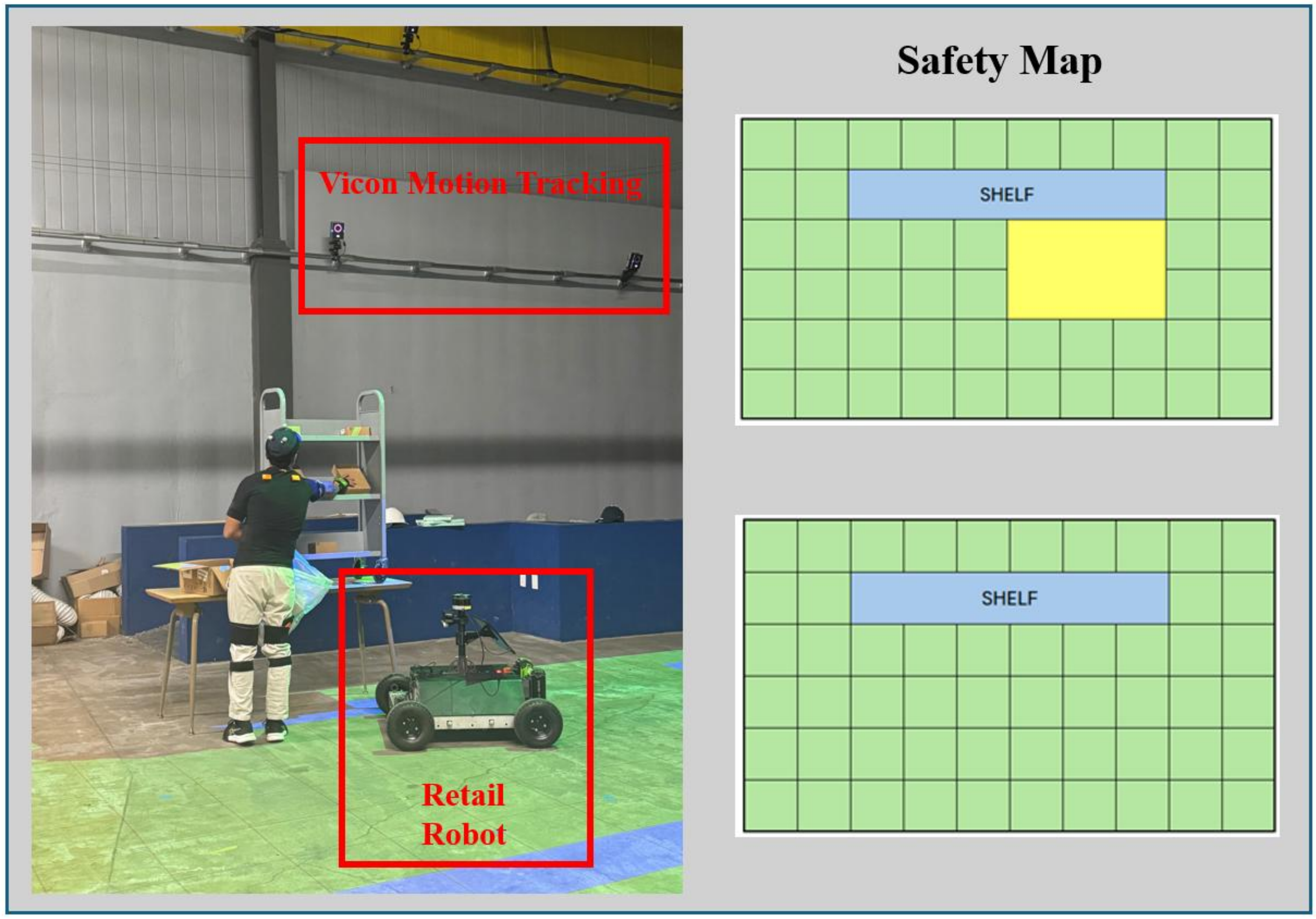

Figure 1. Experimental environment with underlying unsafe regions. Green areas denote safe working regions, yellow areas denote unsafe regions. Bottom right shows participant's view of environment.

The floor was represented as a grid-based map containing shelf locations and predefined safe and unsafe working regions. The unsafe regions represented localized occupational hazards, such as areas contaminated by liquid and therefore associated with increased slipping or falling risk. Although the internal system map distinguished safe and unsafe regions, participants viewed a uniformly green floor projection during all trials. Thus, they could not use visual information to anticipate when they were entering an unsafe region (Figure 1).

A projector displayed the environmental grid in pixel coordinates, whereas the positions of the participant, robot, and shelving unit were tracked in a Vicon world coordinate system expressed in meters. A planar homography transformation mapped locations from the projector coordinate system to the Vicon world coordinate system. This transformation allowed the robot-planning system to relate projected hazard regions to the real-time positions of the participant and robot.

All hazard detection and robot path-planning functions were implemented using Robot Operating System 2 (ROS 2), which supported real-time communication among the motion-tracking system, environmental map, and mobile robot.

### B. Participants

Sixteen healthy adults participated in the data collection. All participants reported being right-hand dominant, being in good general health, and having no history of musculoskeletal disorders during the preceding year. Individuals were eligible if they were adults, could complete the standing and shelf-picking activities independently, and met the health and musculoskeletal criteria described above. Participants provided written informed consent before data collection. The study protocol was approved by the University of Florida Institutional Review Board (IRB No. 202501605).

Kinematic data were available for all 16 participants. EMG data from one participant were excluded because of a data-export error. Therefore, the publicly available EMG data represent 15 participants.

### C. Activities and Tasks

Participants performed a structured retail shelf-picking task under three within-participant experimental conditions: (1) NR, (2) LOW, and (3) HIGH. Each participant completed two trials under each condition, resulting in six trials per participant and 96 expected trials overall.

At the beginning of each trial, the participant stood at a fixed starting location and received a randomized sequence specifying the order in which boxes should be retrieved. Using the right hand, the participant selected the numbered boxes from the shelf and placed them into a shopping bag held in the left hand. This arrangement maintained a consistent asymmetric manual-load component during the task. After retrieving all required boxes, the participant placed the filled shopping bag into the designated collection bin.

In the robot-guided conditions, the AMR monitored the participant's position relative to the predefined unsafe regions. When the participant entered an unsafe region, the robot approached and guided the participant toward a predefined intervention target without verbal instructions or physical contact. After reaching the target, the robot stopped and disengaged, and the participant completed the remaining picking actions (Table 1).

Table 1. Summary of Experimental Activities, Robot-Interaction Conditions, and Data-Collection Coverage

| Activity/condition | Brief operational description | Participants/trials | Typical duration |
|---|---|---|---|
| Calibration and reference activities | Static standing, brief walking, body-measurement collection, and maximum voluntary contraction activities were completed to calibrate the motion-capture and EMG systems. | All eligible participants | Approximately 40 min |
| NR | Participant completed the shelf-picking task without a robot present. | 16 participants; 32 expected trials | Approximately 1 min |
| LOW | The robot approached and guided the participant after entry into an unsafe region. Available linear velocities were 0.2 and 0.3 m/s. | 16 participants; 32 expected trials | Approximately 1 min |
| HIGH | The same robot-guidance procedure was used with available linear velocities of 0.5–0.7 m/s. | 16 participants; 32 expected trials | Approximately 1 min |

For analytical purposes, robot-guided trials could be divided into three phases:

- Pre-interaction phase: From trial onset until the robot entered close proximity to the participant, defined using an approximately 1.5-m threshold.
- Active interaction or transient phase: The interval during which the robot approached and actively guided the participant away from the unsafe region.
- Post-interaction phase: From robot disengagement until completion of the shelf-picking task.

### D. Sensing System

The database contains multimodal measurements of whole-body movement, muscle activity, and human–robot spatial interaction.

**Surface electromyography**

Muscle activity was measured using ten wireless surface EMG sensors from the Trigno Wireless EMG System (Delsys Inc., Boston, MA, USA) (Figure 2(a)). Bilateral measurements were collected from the following muscles:

- Tibialis anterior (R/LTA)
- Lateral gastrocnemius (R/LGAL)
- Rectus femoris (R/LRF)
- Upper trapezius (R/LUT)

The biceps brachii (RBB) and triceps brachii (RTB)were additionally measured on the dominant right arm. EMG signals were normalized using maximum voluntary contraction trials. Raw EMG sampling frequency and exported file format should be reported in the database documentation as .csv files.

**Whole-body inertial motion capture**

Whole-body kinematics were recorded using 17 inertial measurement units from the Xsens Awinda motion-capture system (Xsens Technologies B.V., Enschede, the Netherlands). Sensors were attached according to the manufacturer's full-body configuration. The sensor locations included the head, sternum, pelvis, upper and lower extremities, hands, and feet (Figure 2(b)).

Kinematic data were acquired at 60 Hz and processed using Xsens MVN Analyze/Animate before export. The available kinematic variables include body-segment position, velocity, acceleration, and joint-angle measures. Whole-body center-of-mass variables and measures for selected body segments—including the pelvis, feet, right shoulder, and bilateral knees—were subsequently derived. Movement directions were represented as anteroposterior ($x$), mediolateral ($y$), and vertical ($z$) components relative to the laboratory coordinate system.

**Optical tracking and robot-state data**

A Vicon motion-capture system (Vicon, Oxford, UK) tracked the positions of the participant, mobile robot, and shelving unit within a metric world coordinate system. These measurements supported real-time hazard detection and robot navigation. The robot state included its two-dimensional position, orientation, linear velocity, and angular velocity. Participant location was represented in the same world coordinate system for online planning and collision checking.

The projected hazard map used pixel coordinates. Calibration points visible to both the projector and Vicon systems were used to estimate a homography matrix that transformed the projected coordinates into metric world coordinates.

**Synchronization and calibration**

ROS 2 supported real-time communication among the Vicon tracking system, hazard map, and robot-planning components. Before the experimental trials, the Xsens system was calibrated using a five-second static standing posture followed by a brief walking sequence. This procedure established sensor orientation and supported accurate reconstruction of full-body movement.

The database documentation should additionally state:

- The method used to synchronize Xsens and EMG recordings;
- Whether synchronization was based on a shared hardware trigger, timestamp alignment, or an identifiable event;
- The Vicon and robot-state sampling rates;
- Whether timestamps were retained in the public files; and
- Whether the shared data are raw signals, processed signals, derived outcomes, or a combination of these levels.

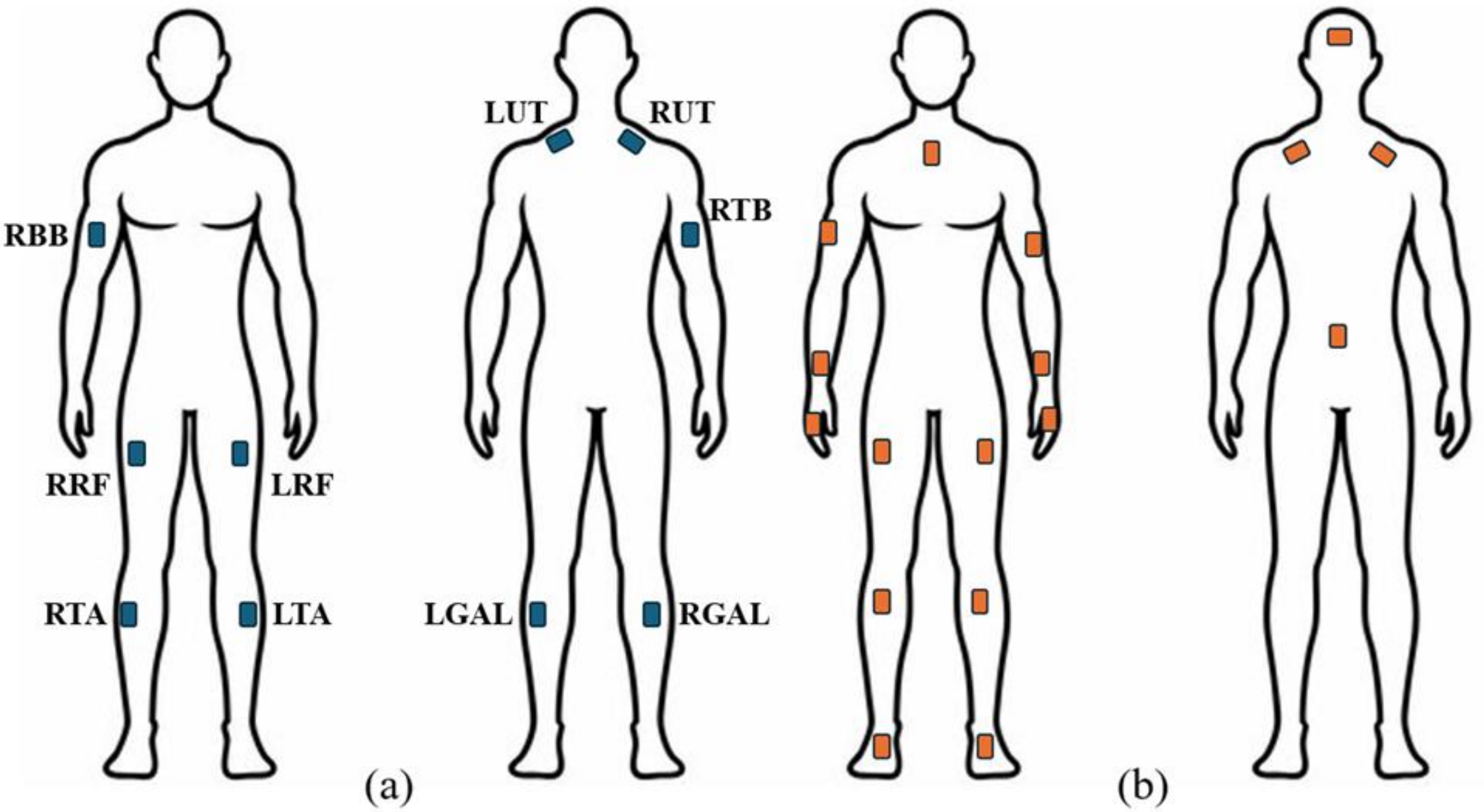

Figure 2. Full body sensor placement for (a) EMG and (b) IMU Xsens sensors

### E. Data Collection Procedure

Each experimental session lasted approximately two hours and followed a standardized sequence:

1. Participant preparation. The participant received an explanation of the study and provided written informed consent. Eligibility and relevant health information were confirmed.
2. Sensor placement. Ten surface EMG sensors were placed over the selected muscles. Seventeen Xsens Awinda sensors were then attached according to the manufacturer's full-body configuration.
3. Anthropometric measurements and calibration. Body measurements required by the motion-capture model were recorded. Xsens calibration consisted of a five-second static standing posture and a brief walking sequence. Maximum voluntary contraction trials were performed for EMG normalization.
4. Task instruction and familiarization. The shelf-picking procedure, shopping-bag requirement, and trial-completion procedure were explained. Participants were not informed of the locations of the predefined unsafe regions.
5. Experimental trials. Each participant completed two trials under each of the NR, LOW, and HIGH conditions. The order of the three conditions was counterbalanced across participants. For every trial, both the required picking sequence and the spatial arrangement of the ten boxes were randomized.
6. Multimodal recording. Kinematic, EMG, spatial-tracking, and robot-state data were recorded while the participant completed the task. In LOW and HIGH trials, entry into an unsafe region triggered the robot-guidance behavior.
7. Trial completion and data storage. A trial ended when the participant retrieved all required boxes and placed the filled shopping bag in the designated collection bin. Files were stored using participant, condition, and repetition identifiers before quality checking and processing.

The experiment generated 96 expected Xsens trials, all of which were retained. Only two isolated sensor-specific Xsens measurements were missing across two trials; the remaining measurements from those trials were preserved. For EMG, 88 of 90 expected trials were available for the 15 participants included in the EMG dataset. Two trials were unavailable because of technical recording or export errors.

Following collection, the Xsens data were processed using the manufacturer's software. EMG data were processed in MATLAB R2023b by removing the DC offset, applying a 60-Hz notch filter, applying a fourth-order Butterworth band-pass filter from 15 to 450 Hz, full-wave rectifying the signals, and applying a fourth-order 6-Hz low-pass filter to obtain the linear envelope. Signals were normalized to the peak value obtained during the maximum voluntary contraction trials. Processed files and associated documentation were subsequently organized for public access through the Ergo-Retail project website.

## IV. Dataset Analysis

The OHRID-Retail database contains multimodal recordings from 16 healthy adults who completed a simulated retail shelf-picking task under three within-participant conditions: NR, LOW, and HIGH. Each participant was scheduled to complete two repetitions of each condition, yielding six trials per participant and 96 planned trials overall. Individual task trials lasted approximately 1 min, and the complete laboratory session, including participant preparation, sensor placement, calibration, and task performance, lasted approximately 2 h.

The release includes whole-body kinematic data recorded using 17 Xsens Awinda inertial sensors at 60 Hz and muscle-activity data recorded using 10 Delsys Trigno surface EMG sensors. The experimental platform also generated Vicon-based spatial tracking and robot-state information for real-time guidance. The public-release description should distinguish clearly between streams used by the experimental system and streams actually deposited in the database (Table 2, Figure 3).

Table 2. Overview of the Ergo-Retail database.

| Characteristic | Database coverage |
|---|---|
| Participants | 16 for kinematics; 15 for EMG |
| Experimental conditions | NR, LOW, and HIGH |
| Repetitions | 2 per participant per condition |
| Planned task trials | 96 Xsens trials; 90 EMG trials after exclusion of one participant's EMG dataset |
| Retained task trials | 96 Xsens trials; 88 EMG trials |
| Typical trial duration | Approximately 1 min |
| Typical session duration | Approximately 2 h, including preparation and calibration |
| Kinematic sensing | 17 Xsens Awinda IMUs sampled at 60 Hz |
| Muscle-activity sensing | 10 Delsys Trigno surface EMG sensors |
| Additional experimental streams | Vicon spatial tracking and robot-state information; inclusion in the public release should be confirmed |

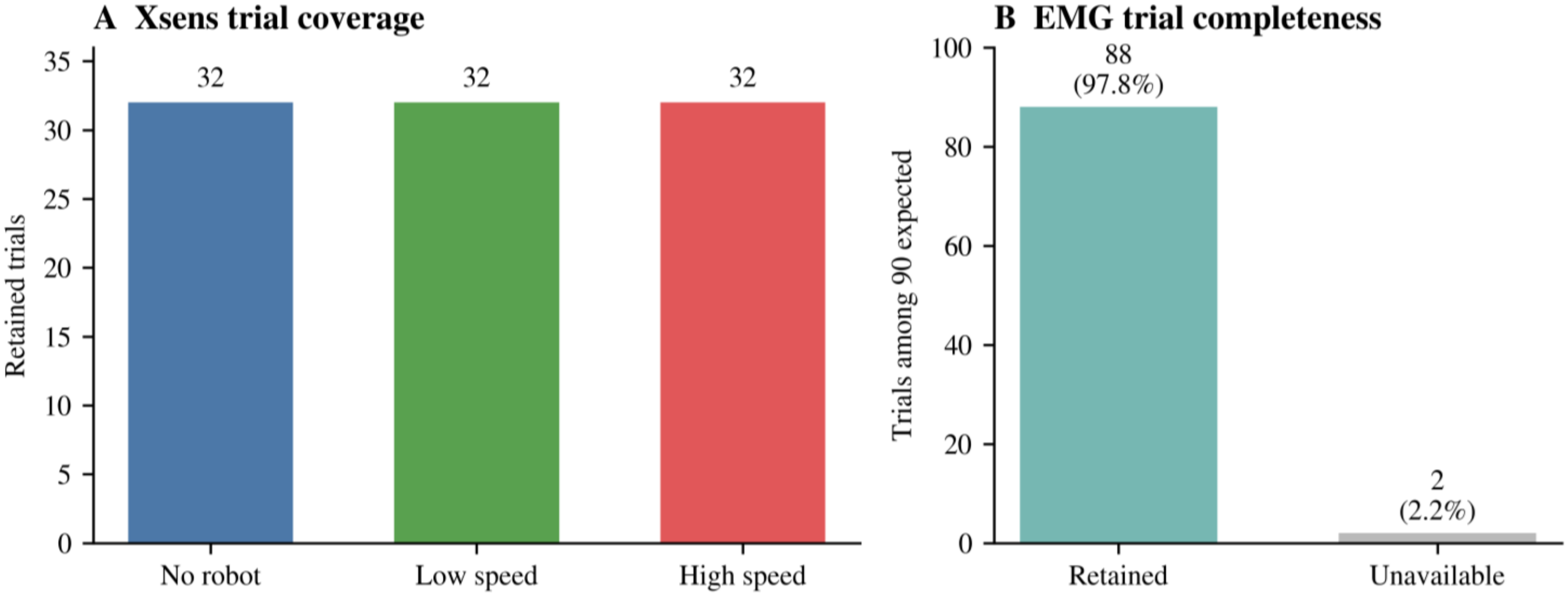

Figure 3. Dataset coverage by sensing modality

The released Xsens component contains 96 usable trial recordings from 16 participants. At 60 Hz, the Xsens dataset contains 224,613 temporal data points, representing 1.04 h (62.4 min) of cumulative recording time. All expected Xsens recordings were available.The EMG component contains 88 usable trial recordings from 15 participants. At a sampling frequency of approximately 2148.148 Hz, the EMG dataset contains 8,003,448 temporal data points, representing approximately 1.03 h (62.1 min) of cumulative recording time.

### B. Participant Distribution

The kinematic component includes 16 participants, whereas the EMG component includes 15 participants. Participants had a mean age of 25.4 years (SD = 3.0), mean height of 176.8 cm (SD = 5.8), and mean body mass of 64.1 kg (SD = 6.4). All participants were right-hand dominant, reported good general health, and reported no musculoskeletal disorder during the preceding year.

### C. Activity Distribution

All participants performed the same shelf-picking activity under the three robot-interaction conditions. The NR condition contained 32 retained Xsens trials, and the LOW and HIGH conditions each contained 32 retained Xsens trials. Thus, the kinematic dataset is balanced across the three conditions. The two LOW trials used robot linear velocities selected from 0.2 and 0.3 m/s, whereas the two HIGH trials used velocities selected from 0.5, 0.6, and 0.7 m/s. The order of the three conditions was counterbalanced, and the box arrangement and picking sequence were randomized for each trial.

### D. Movement and Sensor Characteristics

Descriptive summaries illustrate the range of whole-body movement and muscle activation represented in the database. Across the complete task, mean center-of-mass (CoM) velocity magnitude was 0.04 ± 0.01 m/s under NR, 0.07 ± 0.04 m/s under LOW, and 0.07 ± 0.03 m/s under HIGH. Mean CoM acceleration magnitude was 0.19 ± 0.04 m/s² under NR, 0.29 ± 0.11 m/s² under LOW, and 0.28 ± 0.10 m/s² under HIGH. These descriptive values demonstrate that the robot-guided recordings contain greater overall movement intensity than the no-robot recordings (Figure 4).

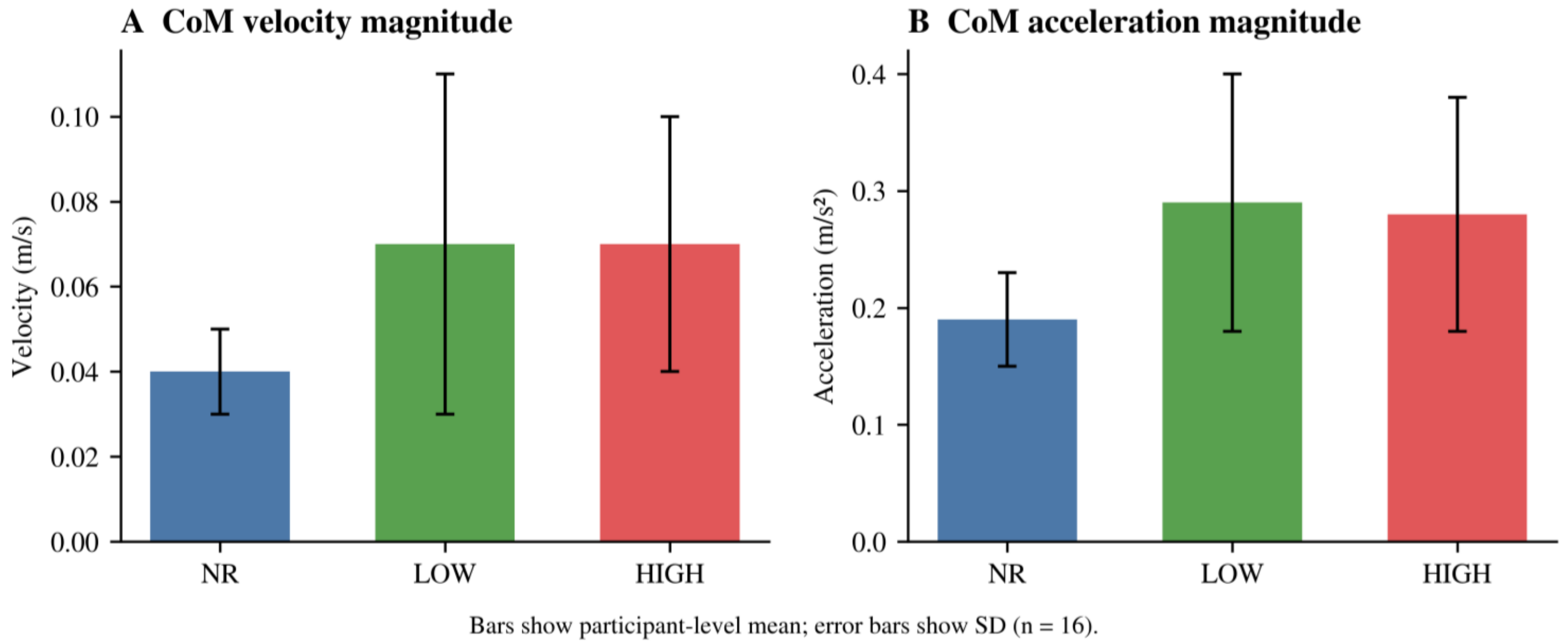

Figure 4. Representative whole body movement characteristics

The EMG component contains bilateral tibialis anterior, lateral gastrocnemius, rectus femoris, and upper trapezius measurements, together with right-arm biceps brachii and triceps brachii measurements. Figure 5 presents participant-level mean integrated EMG values across all ten sensor locations and three conditions. The displayed variation across body locations illustrates the diversity of neuromuscular activity captured during lower-limb postural control, upper-body reaching, and asymmetric bag handling.

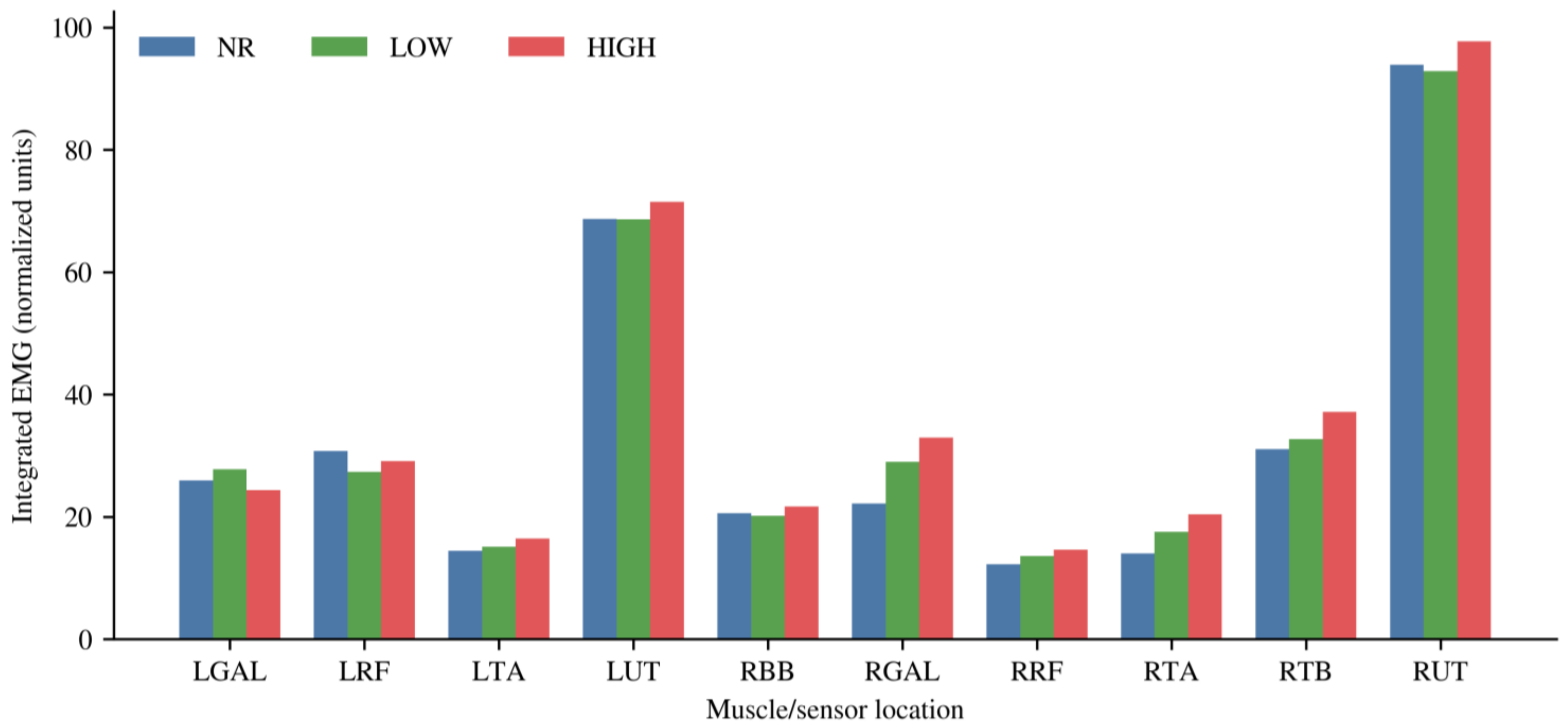

Figure 5. Integrated EMG profiles across task conditions

The available notebook also derives transient-phase CoM sample entropy, SPARC movement smoothness, pelvis-to-foot acceleration redistribution, EMG redistribution, and mediolateral margin of stability. These processed features may be listed as derived variables in the database documentation.

- Provide simple descriptive examples showing how recorded movement/sensor signals vary across activities and body locations.
- Possible summaries include movement intensity, acceleration magnitude, activity duration, representative time series, or other already-available features.
- The purpose is to demonstrate the range and diversity of the data, not to establish a new scientific hypothesis or benchmark.

## V. Discussion

OHRID-Retail provides an open, human-centered multimodal dataset designed to characterize human behavior in a structured retail environment with varying levels of mobile-robot assistance. Whereas many existing robotics datasets primarily emphasize robot observations, manipulation, and task execution, the present dataset focuses on the human side of shared environments. By combining synchronized full-body movement and physiological measurements, OHRID-Retail provides a complementary resource for examining how people move, perform physical activities, and respond to changes in their surrounding environment. This human-centered perspective is particularly relevant as mobile robots become increasingly common in retail and other shared spaces, where successful deployment requires not only effective robot navigation but also an understanding of the people with whom robots share the environment.

The dataset may support several categories of future research. At the human-performance level, the movement and physiological recordings can be used to characterize task execution, movement strategies, and behavioral adaptation during retail activities. The synchronized multimodal signals may also support development and evaluation of human activity recognition, movement classification, and multimodal sensor-fusion methods. From an ergonomics perspective, the data provide opportunities to examine movement patterns and physiological responses associated with shelf picking, object handling, walking, and robot-assisted task performance. From a robotics perspective, human trajectories and behavioral responses may contribute to the development of human-aware navigation and shared-space models in which robot behavior is informed by how people move and respond to nearby robotic systems. These applications are not mutually exclusive, and the open structure of the dataset is intended to allow researchers to formulate questions beyond those considered in the original data collection.

OHRID-Retail should therefore be viewed as complementary to, rather than a replacement for, existing robot-centered and HRI datasets. Large-scale resources such as DROID provide extensive information about robot actions and interactions with physical environments, whereas datasets such as HARMONIC and Kaiwu capture multimodal information in specific human–robot collaborative tasks. OHRID-Retail contributes a different perspective by emphasizing detailed human movement and physiological data within a retail-relevant shared environment. The present release does not attempt to represent the complete human–robot system or the full range of interactions encountered in operational retail settings. Instead, it provides a structured human-side data resource that can be combined with and inform future robot-centered data collection.

Several limitations should be considered when using OHRID-Retail. First, data were collected in a controlled retail-like environment rather than an operating retail store. Although this design improves experimental repeatability, it does not capture the full variability of naturalistic retail environments, including pedestrian traffic, environmental distractions, and interactions with other people. Second, the current sample includes only 16 healthy young adults, limiting representation of age-related and functional diversity. Third, although the protocol includes an autonomous mobile robot and different guidance speeds, the interaction is limited to a specific robot-guidance scenario. The current release therefore represents a structured subset of the broader range of human–robot interactions that may occur in retail environments.

Future development of OHRID-Retail can address these limitations by expanding to larger and more diverse participant populations, additional retail activities and environmental configurations, and a wider range of human–robot encounters. Additional sensing modalities and increasingly naturalistic environments could further connect human movement and physiological responses with robot behavior, environmental context, and interaction events. Such extensions would progressively expand the current human-centered dataset toward a broader multimodal resource for studying human–robot coexistence.

Overall, OHRID-Retail is intended as an openly accessible and reusable foundation rather than a comprehensive representation of retail human–robot interaction. The current release provides a structured human-side resource for research in human movement, ergonomics, activity recognition, human-aware robotics, and HRI, while establishing a framework that can be expanded in future releases.

## VI. Conclusion

This paper introduces OHRID-Retail, an open multimodal dataset of human activity collected in a structured retail environment under varying levels of mobile-robot assistance. By combining synchronized human movement and physiological measurements with standardized retail tasks and controlled robot-interaction conditions, the dataset provides a human-centered complement to existing robot-focused data resources. OHRID-Retail is intended to support research in human movement, ergonomics, activity recognition, human-aware robotics, and human–robot interaction, while providing a common framework for future data expansion. As the initiative grows to incorporate more diverse participants, activities, environments, and human–robot encounters, OHRID-Retail can serve as an evolving open resource for understanding human behavior in increasingly shared human–robot environments.

## References

Ciliberto, M., Fortes Rey, V., Calatroni, A., Lukowicz, P., & Roggen, D. (2021). Opportunity++: A multimodal dataset for video- and wearable, object and ambient sensors-based human activity recognition. *Frontiers in Computer Science, 3*, Article 792065. https://doi.org/10.3389/fcomp.2021.792065

DelPreto, J., Liu, C., Luo, Y., Foshey, M., Li, Y., Torralba, A., Matusik, W., & Rus, D. (2022). ActionSense: A multimodal dataset and recording framework for human activities using wearable sensors in a kitchen environment. *Advances in Neural Information Processing Systems, 35*, 13800–13813. https://proceedings.neurips.cc/paper_files/paper/2022/file/5985e81d65605827ac35401999aea22a-Paper-Datasets_and_Benchmarks.pdf

Ghorbani, S., Mahdaviani, K., Thaler, A., Kording, K., Cook, D. J., Blohm, G., & Troje, N. F. (2021). MoVi: A large multi-purpose human motion and video dataset. *PLOS ONE, 16*(6), Article e0253157. https://doi.org/10.1371/journal.pone.0253157

Grauman, K., Westbury, A., Byrne, E., Chavis, Z., Furnari, A., Girdhar, R., Hamburger, J., Jiang, H., Liu, M., Liu, X., Martin, M., Nagarajan, T., Radosavovic, I., Ramakrishnan, S. K., Ryan, F., Sharma, J., Wray, M., Xu, M., Xu, E. Z., ... Malik, J. (2022). Ego4D: Around the world in 3,000 hours of egocentric video. In *Proceedings of the IEEE/CVF Conference on Computer Vision and Pattern Recognition* (pp. 18995–19012). https://openaccess.thecvf.com/content/CVPR2022/html/Grauman_Ego4D_Around_the_World_in_3000_Hours_of_Egocentric_Video_CVPR_2022_paper.html

Iodice, F., De Momi, E., & Ajoudani, A. (2022). HRI30: An action recognition dataset for industrial human-robot interaction. In *2022 26th International Conference on Pattern Recognition (ICPR)* (pp. 4941–4947). IEEE. https://doi.org/10.1109/ICPR56361.2022.9956300

Jiang, S., Li, H., Ren, R., Zhou, Y., Wang, Z., & He, B. (2025). Kaiwu: A multimodal manipulation dataset and framework for robot learning and human-robot interaction. *IEEE Robotics and Automation Letters, 10*(11), 11482–11489. https://doi.org/10.1109/LRA.2025.3609615

Khazatsky, A., Pertsch, K., Nair, S., Balakrishna, A., Dasari, S., Karamcheti, S., Nasiriany, S., Srirama, M. K., Chen, L. Y., Ellis, K., Fagan, P. D., Hejna, J., Itkina, M., Lepert, M., Ma, Y. J., Miller, P. T., Wu, J., Belkhale, S., Dass, S., ... Finn, C. (2024). DROID: A large-scale in-the-wild robot manipulation dataset. In *Robotics: Science and Systems XX*. https://doi.org/10.15607/RSS.2024.XX.120

Kong, Q., Wu, Z., Deng, Z., Klinkigt, M., Tong, B., & Murakami, T. (2019). MMAct: A large-scale dataset for cross modal human action understanding. In *Proceedings of the IEEE/CVF International Conference on Computer Vision* (pp. 8657–8666). https://doi.org/10.1109/ICCV.2019.00875

Newman, B. A., Aronson, R. M., Srinivasa, S. S., Kitani, K., & Admoni, H. (2022). HARMONIC: A multimodal dataset of assistive human-robot collaboration. *The International Journal of Robotics Research, 41*(1), 3–11. https://doi.org/10.1177/02783649211050677

Reiss, A., & Stricker, D. (2012). Introducing a new benchmarked dataset for activity monitoring. In *2012 16th International Symposium on Wearable Computers* (pp. 108–109). IEEE. https://doi.org/10.1109/ISWC.2012.13

Shahroudy, A., Liu, J., Ng, T.-T., & Wang, G. (2016). NTU RGB+D: A large scale dataset for 3D human activity analysis. In *Proceedings of the IEEE Conference on Computer Vision and Pattern Recognition* (pp. 1010–1019). https://openaccess.thecvf.com/content_cvpr_2016/html/Shahroudy_NTU_RGBD_A_CVPR_2016_paper.html

Sigal, L., Balan, A. O., & Black, M. J. (2010). HumanEva: Synchronized video and motion capture dataset and baseline algorithm for evaluation of articulated human motion. *International Journal of Computer Vision, 87*(1–2), 4–27. https://doi.org/10.1007/s11263-009-0273-6

Yan, M., Zhang, Y., Cai, S., Fan, S., Lin, X., Dai, Y., Shen, S., Wen, C., Xu, L., Ma, Y., & Wang, C. (2024). RELI11D: A comprehensive multimodal human motion dataset and method. In *Proceedings of the IEEE/CVF Conference on Computer Vision and Pattern Recognition* (pp. 2250–2262). https://openaccess.thecvf.com/content/CVPR2024/html/Yan_RELI11D_A_Comprehensive_Multimodal_Human_Motion_Dataset_and_Method_CVPR_2024_paper.html

Zhang, Y., Lin, J., Zeng, A., Wu, G., Lu, S., Fu, Y., Cai, Y., Zhang, R., Wang, H., & Zhang, L. (2025). *Motion-X++: A large-scale multimodal 3D whole-body human motion dataset* [Preprint]. arXiv. https://doi.org/10.48550/arXiv.2501.05098